\documentclass[times]{sagej}

\IfFileExists{latexml.sty}{%
  \usepackage{latexml}%
}{%
  \newif\iflatexml
  \latexmlfalse
}

\usepackage[
  papersize={210mm,280mm},
  text={170.5mm,226mm},
  headsep=21pt,
  footskip=18pt,
  centering
]{geometry}

\usepackage[T1]{fontenc}
\usepackage[utf8]{inputenc}
\usepackage[english]{babel}

\usepackage{amsmath}
\usepackage{graphicx}
\usepackage{multirow}
\usepackage{array}
\usepackage{booktabs}
\usepackage{tabularx}

\usepackage[numbers,square,comma,sort&compress]{natbib}

\renewcommand{\refname}{{\normalsize References}}

\usepackage[hidelinks]{hyperref}
\usepackage{microtype}

\renewcommand{\footnoterule}{%
  \kern-3pt
  \hrule width 40mm
  \kern 3mm
}

\newcommand{\submitteddate}{7 September 2025}
\newcommand{\accepteddate}{7 September 2026}

\renewcommand{\journalname}{Journal of Information Science}

\makeatletter

\renewcommand{\@seccntformat}[1]{%
  \csname the#1\endcsname.\hspace{0.5em}%
}

\def\ps@title{%
  \def\@oddhead{}%
  \let\@evenhead\@oddhead
  \def\@oddfoot{}%
  \let\@evenfoot\@oddfoot
}

\def\ps@sagepage{%
  \let\@mkboth\@gobbletwo

  \def\@oddhead{%
    \parbox{\textwidth}{%
      \normalsize\sagesf\itshape
      \leftmark
      \hfill
      \upshape\thepage\\[-6pt]
      \noindent\rule{\textwidth}{0.25pt}%
    }%
  }%

  \let\@evenhead\@oddhead

  \def\@oddfoot{}%
  \let\@evenfoot\@oddfoot
}

\def\@maketitle{%

  \vspace*{-30pt}%
  \null%

  {\noindent
   \sagesf
   \normalsize
   \itshape
   Article
   \par}

  \vspace{4pt}

  \noindent\rule{\textwidth}{0.5pt}

  \vspace{14pt}

  \noindent
  \begin{minipage}[t]{\dimexpr\textwidth-57.625mm\relax}

    \raggedright
    \sagesf
    \titlesize
    \bfseries
    \@title
    \par

  \end{minipage}%
  \hspace{15mm}%
  \begin{minipage}[t]{42.625mm}

    \raggedright
    \sagesf
    \scriptsize

    \journalname\\[3pt]
    First submitted: \submitteddate\\
    Accepted: \accepteddate\\
    \href{https://doi.org/10.1177/01655515261490074}
         {\textcolor{blue}{View journal version}}

  \end{minipage}

  \par

  \vspace{25pt}

  {\noindent
   \raggedright
   \sagesf
   \large
   \bfseries
   \@author
   \par}

  \vspace{30pt}

  {\noindent
   \usebox\absbox
   \par}

  \vspace{17pt}

  {\noindent
   \@keywords
   \par}

  \vspace{20pt}
}

\renewcommand\maketitle{\par

  \begingroup

    \if@twocolumn

      \ifnum\col@number=\@ne

        \@maketitle

      \else

        \twocolumn[\@maketitle]%

      \fi

    \else

      \global\@topnum\z@%

      \@maketitle

    \fi

    \thispagestyle{title}%
    \label{FirstPage}%
    \@affiliation
    \@corrauth
    \@email

  \endgroup

  \global\let\affiliation\relax
  \global\let\thanks\relax
  \global\let\maketitle\relax
  \global\let\@maketitle\relax
  \global\let\@thanks\@empty
  \global\let\@author\@empty
  \global\let\@date\@empty
  \global\let\@title\@empty
  \global\let\@affiliation\@empty
  \global\let\title\relax
  \global\let\author\relax
  \global\let\date\relax
  \global\let\and\relax
}

\makeatother

\newcommand{\papertitle}{%
Multi-functional embedding models for funder name
disambiguation in scientific publication records%
}

\newcommand{\paperabstract}{%
Understanding the historical allocation and distribution of research funding advances our knowledge of how scientific research is supported across fields, institutions, and regions. However, large-scale analyses are hindered by the lack of comprehensive funder name disambiguation solutions, as funder names often exhibit spelling variations, translations, abbreviations, and inconsistent levels of granularity. In this paper, we present a framework for developing multilingual, multi-functional funder name disambiguation models and demonstrate its application to research publications in biodiversity conservation. To construct a training dataset, we integrated the Research Organization Registry (ROR), which provides unique identifiers for research organizations, with two publication datasets: the Web of Science (WoS) and the Crossref Open Funder Registry (OFR). We used multi-task learning with Contrastive Loss and Multiple Negatives Ranking Loss to fine-tune three open-weight embedding models from the Sentence Transformer, Gemma, and Qwen3 families. The best-performing models achieved accuracy above 0.90 when matching WoS funder names to ROR identifiers, outperforming general-purpose LLMs, including GPT-5.2, Claude-Sonnet-4.6, and Gemini-2.5-Flash, by more than 0.1. For funder names not indexed in ROR, we constructed a similarity network among funder names and identified clusters within it. Finally, we analyzed the disambiguation results and highlighted challenges arising from limited knowledge of smaller funders and funders from non-English-speaking countries. This work provides a reusable framework for funder name disambiguation with potential applicability across different model architectures and datasets, featuring cost-effective training data creation and multi-task learning and disambiguation.%
}

\newcommand{\paperkeywords}{%
funder name disambiguation;
publication records;
biodiversity conservation;
multi-task learning;
embedding models%
}

\iflatexml

\else

\title{\papertitle}

\author{%
  Kanyao Han\affilnum{1},
  Zhiwen You\affilnum{1},
  Jinseok Kim\affilnum{2} and
  Jana Diesner\affilnum{1,3,4,5}%
}

\affiliation{%
  \affilnum{1} University of Illinois at Urbana-Champaign, Champaign, USA\\
  \affilnum{2} University of Michigan, Ann Arbor, USA\\
  \affilnum{3} Technical University of Munich, Munich, Germany\\
  \affilnum{4} Munich Center for Machine Learning, Munich, Germany\\
  \affilnum{5} Munich Data Science Institute, Munich, Germany%
}

\corrauth{%
  Kanyao Han,
  University of Illinois at Urbana-Champaign,
  Champaign, USA\\
  Email: kanyaoh2@gmail.com
}

\runninghead{Han et al.}

\fi

\begin{document}

\iflatexml

%
%
%
%
%
%


\begin{center}

{\LARGE\bfseries
Multi-Functional Embedding Models for Funder Name\\
Disambiguation in Scientific Publication Records%
\footnote{%
\begingroup
\LARGE\normalfont
\renewcommand{\arraystretch}{0.72}

\begin{tabular}{@{}l@{}}

\textbf{Journal of Information Science}\\[-0.25em]
First submitted: \submitteddate\\[-0.25em]
Accepted: \accepteddate\\[-0.25em]
\href{https://doi.org/10.1177/01655515261490074}
{View journal version}\\[0.8em]

\textbf{Corresponding author:}\\[-0.25em]
Kanyao Han\\[-0.25em]
Email: \href{mailto:kanyaoh2@gmail.com}{kanyaoh2@gmail.com}

\end{tabular}

\endgroup
}%
\par}

\end{center}

\bigskip


\begin{center}

{\large
Kanyao Han\textsuperscript{1},
Zhiwen You\textsuperscript{1},
Jinseok Kim\textsuperscript{2} and
Jana Diesner\textsuperscript{1,3,4,5}
\par}

\end{center}

\medskip


\begin{center}

{\small

\textsuperscript{1}
University of Illinois at Urbana-Champaign,
Champaign, USA\\

\textsuperscript{2}
University of Michigan,
Ann Arbor, USA\\

\textsuperscript{3}
Technical University of Munich,
Munich, Germany\\

\textsuperscript{4}
Munich Center for Machine Learning,
Munich, Germany\\

\textsuperscript{5}
Munich Data Science Institute,
Munich, Germany

\par}

\end{center}

\bigskip

%

\section*{Abstract}

\paperabstract

\bigskip

%

\noindent
\textbf{Keywords}

\paperkeywords

\bigskip

\else


%

\begin{abstract}

\paperabstract

\end{abstract}

%



\maketitle


\setcounter{footnote}{0}
\renewcommand{\thefootnote}{\arabic{footnote}}

\fi

\section{Introduction} \label{dism}
Research funding is essential to advance science and scholarship \citep{whitley2018impact,zhou2020depth}. Analyses of funding allocation to individuals, sociodemographic groups, fields, and organizations can improve our understanding of issues in funding acquisition or provision. A considerable body of research has examined funding allocation practices \citep{bromham2016interdisciplinary,xie2014undemocracy,waldron2013targeting,zhi2016}. Much of this research examines the funding allocation practices of one or a few prestigious funders, such as the National Science Foundation (NSF) \citep{rabovsky2014higher} and the National Institutes of Health (NIH) \citep{jacob2011impact,ballreich2021allocation} in the United States, the National Natural Science Foundation of China (NSFC) \citep{zhi2016}, or the Natural Sciences and Engineering Research Council of Canada (NSERC) \citep{fortin2013big}. For instance, Zhi and Meng \citep{zhi2016} analyzed how the NSFC distributed its funds across institutions, cities, and life science fields, uncovering significant disparities. Despite the large body of literature, large-scale comparative analyses across multiple funders and countries remain scarce, primarily due to the lack of accessible, comprehensive, and high-quality funding data \citep{bloch2015size}.

One major obstacle to funding data analysis is the lack of comprehensive mappings that link funder name occurrences, including variations in spelling, translations, and abbreviations, to standardized records of unique funder identifiers \citep{lammey2020solutions,waldron2017reductions}. Many bibliometric data providers or sources, such as the Web of Science (WoS), Scopus, and PubMed, record funder names for some of the papers they index. However, the coverage and quality of these funder name records vary across data sources \citep{kokol2018discrepancies,liu2020accuracy,paul2016characterization}. For example, prior studies \citep{kokol2018discrepancies,liu2020accuracy} found that WoS has the highest coverage of papers with funder name records (i.e., a list of funder names per paper). Funder names indexed in WoS are mainly extracted from the acknowledgment sections of papers \citep{pranckute2021web}, such that the same funder might be referred to by different names. For instance, the National Science Foundation, a major science funder based in the United States, may be recorded as ``NSF'', ``National Science Foundation'', ``the U.S. National Science Foundation'', among other variations. This issue is further complicated by translations, name changes, misspellings, and the level of resolution for funders (e.g., NSF Graduate Research Fellowship). Recently, academic communities have been curating cleaner versions of funder names by integrating publishing industry data and using crowdsourcing approaches, such as through the Crossref Open Funder Registry (OFR). The OFR, originally contributed by Elsevier and now maintained by Crossref, allows authors to report funder names of their papers and select a unique OFR funder ID for each reported funder name. Existing entries are also reviewed to ensure accuracy \citep{hendricks2020crossref}. However, OFR has lower coverage of papers with funder records, for example, in the domain of biodiversity conservation. Specifically, we found that 122,508 papers categorized as biodiversity conservation have been published since 1900 in the WoS corpus. Of those, 52,760 papers have funder name records in the WoS data. In the OFR data, the number of papers with funder records is 20,793.

To address the gap between the lack of high-quality, high-coverage funder records and the data needed for funding-allocation studies, we leverage multiple funder data sources to develop a funder disambiguation model for biodiversity conservation. Because research in this domain is supported by organizations and programs worldwide across both the natural and social sciences, it captures much of the complexity of funder naming and provides a useful testbed for a broadly applicable model.

Name disambiguation is an established NLP task, with a large body of literature dedicated to designing and discussing methods for resolving named entities \citep{ali2022named, bouarroudj2022named} to advance studies in the science of science \citep{diesner2015impact,kim2014name,kim2016distortive,kim2015effect,mishra2018self}, among other fields. Although much is known about the impact of name ambiguity \citep{kim2016distortive} and how to disambiguate author names \citep{sanyal2021review, zhang2023lagos}, little attention has been paid to the disambiguation of funder names. In addition to studies related to author name disambiguation, there are a few prior studies on institution name disambiguation \citep{ancona2023novel,jonnalagadda2010nemo,shao2020elad,song2020research}. However, these studies on institution name disambiguation focus on author affiliation names rather than funder names. Specifically, although there is a large overlap between author affiliations and funders as the main task for both is to disambiguate certain entity names\footnote{This paper uses three related terms: \textbf{organization}, \textbf{institution}, and \textbf{funder}. \textbf{Organization} refers broadly to an organized entity, such as a research organization, government agency, or company, while \textbf{institution}, especially in bibliometric research, is commonly used to refer to an academic or research organization with which an author of a paper is affiliated. \textbf{Funder} refers to an entity reported as providing research funding and does not necessarily correspond to an organization or institution, as researchers may also report funding programs or even individuals as funders.}, funder names include more non-research organizations (such as governmental departments, non-governmental organizations, and small businesses) because authors of scholarly papers are mainly from research organizations. Furthermore, most of these prior studies \citep{ancona2023novel,jonnalagadda2010nemo} tried to identify shared components in institution names, including but not limited to characters, words, and n-grams, as well as the sequence of these components. These approaches often fail when a name disambiguation task requires external knowledge. For example, the abbreviated name ``MacArthur Foundation'' is often used to refer to the ``John D. and Catherine T. MacArthur Foundation'' in the U.S., rather than the ``Ellen MacArthur Foundation'', a distinct organization based in the U.K. We need such external and contextualizing knowledge to identify which funder an abbreviated name such as ``MacArthur Foundation'' refers to rather than relying solely on shared components. Therefore, fine-tuning pre-trained models using funder name data offers an alternative approach that leverages knowledge acquired through large-scale pre-training while incorporating domain-specific knowledge during fine-tuning \citep{roberts2020much}.

\begin{figure}[!htb]
    \centering
    \includegraphics[width=1\textwidth]{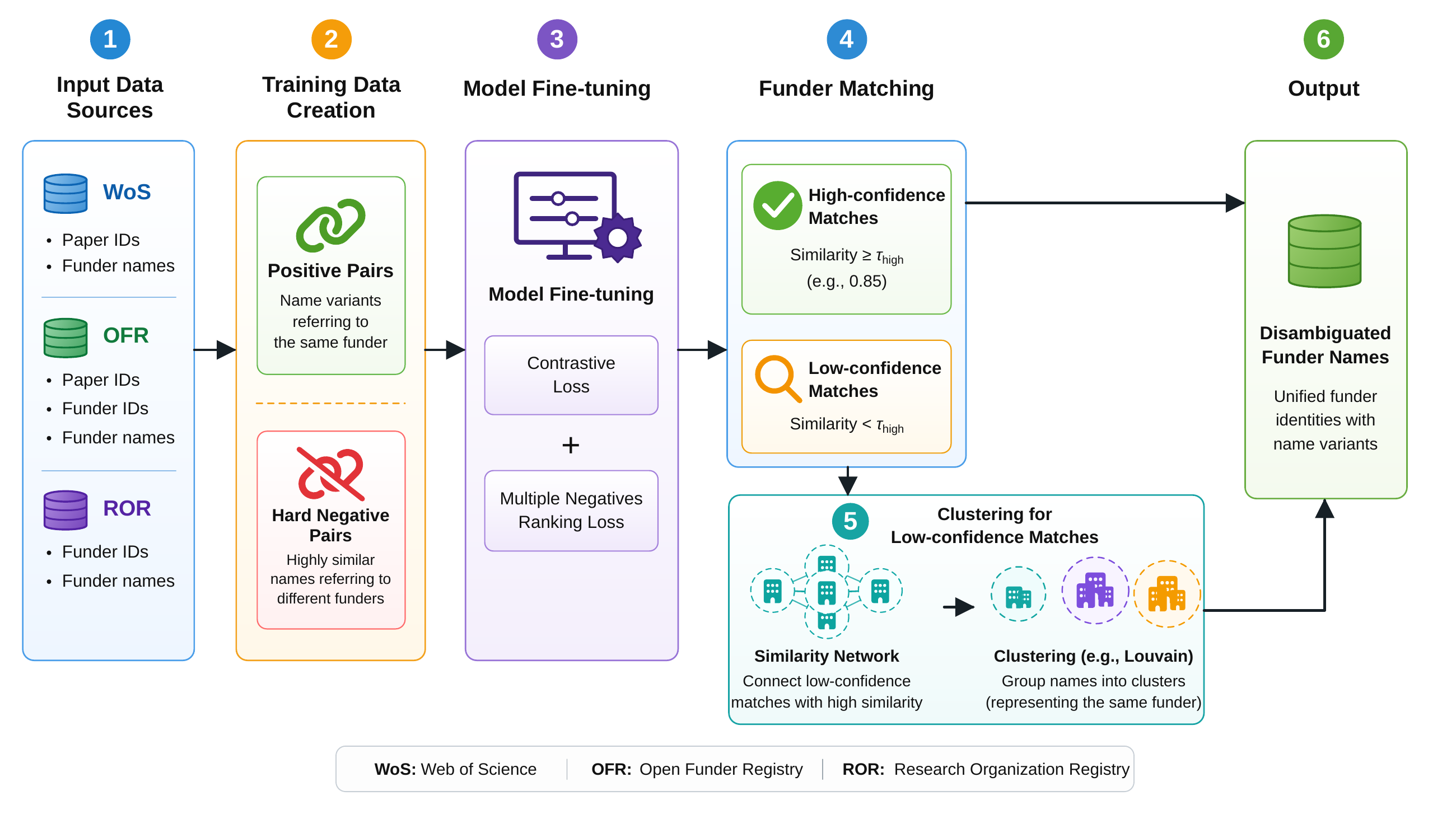}
    \caption{Overall funder name disambiguation workflow}
    \label{fig:simworkflow}
\end{figure}

Figure~\ref{fig:simworkflow} presents our workflow for model fine-tuning and the overall funder name disambiguation process. The workflow begins with the creation of a training dataset to address the limited availability of labeled funder-name pairs. Fine-tuning state-of-the-art embedding models typically requires training data in the format of [text1, text2, label] \citep{reimers-2019-sentence-bert}, where each pair of funder names is accompanied by a label indicating whether the two name strings refer to the same funder. To construct such a dataset, we identified papers that are indexed by both WoS and OFR and contain funder name records. We constructed pairs of funder names per paper (i.e., text1 = funder name in WoS, text2 = funder name in OFR) that refer to the same funder but are recorded differently in the two sources, based on similarity scores computed by a pre-trained Sentence Transformer (ST) model \citep{reimers-2019-sentence-bert}. We then replaced the names derived from OFR in each pair with the funder name from the Research Organization Registry (ROR), a dataset that maps name variants of funders to shared unique funder IDs with OFR, for data augmentation. This data augmentation step ensures that the model learns as many name variants of the same funder as possible. We used the resulting dataset to fine-tune the pre-trained ST \citep{reimers-2019-sentence-bert}, Gemma \citep{vera2025embeddinggemma}, and Qwen3 \citep{zhang2025qwen3} embedding models using two loss functions: Contrastive Loss, which classifies whether two funder names refer to the same funder or not, and Multiple Negatives Ranking Loss, which ranks candidate names based on their likelihood of referring to the same funder. Among the funders with unique funder IDs documented in the ROR data, the fine-tuned ST and Gemma embedding models achieved accuracies of 0.98 for names occurring in ROR and over 0.91 for names occurring in WoS, outperforming GPT-5.2\footnote{\url{https://openai.com/index/introducing-gpt-5-2/}}, Claude-Sonnet-4.6\footnote{\url{https://www.anthropic.com/news/claude-sonnet-4-6}}, and Gemini-2.5-Flash\footnote{\url{https://ai.google.dev/gemini-api/docs/models/gemini-2.5-flash}} by more than 0.1 ($p < .001$ for all pairwise comparisons based on two-sided McNemar tests).

Using the fine-tuned ST model (with an F1 score of 0.92) for classification, we identified approximately 37.4\% of funder names in WoS that were not indexed in ROR data and thus cannot be assigned an existing unique funder ID. To address this gap, we constructed a similarity network by linking highly similar funder names using the similarity scores computed by our disambiguation model. We then applied the Louvain algorithm \citep{blondel2008fast} to cluster these funder names to identify and disambiguate major funders in biodiversity conservation research, particularly those from non-English-speaking countries, such as the ``National Basic Research Program (973 Program).'' This example also highlights linguistic differences in how the term ``funder'' is defined and referenced. For instance, in China, the term ``program'' is often used to refer to funders, even when the official name of the organization should be, for example, the ``Office of the 973 Program.''

With this work, we make four contributions:

\begin{itemize}
    \item First, we introduce a comprehensive and reusable framework for funder name disambiguation, covering training data creation, model fine-tuning, evaluation, and application. The framework supports multiple disambiguation tasks, including determining whether two names refer to the same funder or organization, matching funder names against a list of candidates, and clustering funder names. It is also compatible with diverse model architectures, including Sentence Transformer, Gemma, and Qwen.

    \item Second, through extensive evaluation, we demonstrate that fine-tuned embedding models using our framework substantially improve funder name disambiguation compared with pre-trained embedding models and generative LLMs. Although the models are trained using data from biodiversity conservation publications, the training data include funders operating across research domains, suggesting the potential applicability of the approach beyond biodiversity conservation.

    \item Third, we construct a large-scale disambiguated funder dataset for biodiversity conservation publications by integrating WoS funder records with publicly available organizational information. Due to licensing restrictions on the underlying WoS data, the disambiguated dataset and fine-tuned embedding models cannot be publicly released\footnote{The disambiguated dataset and fine-tuned embedding models may be shared with researchers who have appropriate access to the underlying licensed WoS data. Researchers with access to the underlying WoS data may contact the authors and Clarivate to inquire about access to the disambiguated dataset and fine-tuned models.}. To facilitate reproducibility, we provide the publication identifiers (WoS IDs, DOIs, and paper titles), code, and documentation needed to reconstruct the dataset and reproduce the model training (see Section \ref{data}).

    \item Fourth, we characterize and detail the challenges in funder name disambiguation, such as the lack of comprehensive funder records from non-English-speaking countries and smaller funders. These findings highlight limitations in existing funder registries and bibliometric data that should be considered in large-scale funding analyses.

\end{itemize}

\section{Related Work}
Funder name disambiguation is a special case of name disambiguation and is most similar to the task of institution (i.e., author affiliation) disambiguation in scholarly publications. We refer to the latter task as institution name disambiguation in this paper. An example institution name is ``Department of Statistics, Virginia Tech, Blacksburg, USA, address@vt.edu.''

There are generally two approaches to institution name disambiguation \citep{huang2014institution,shao2020elad}: 1) matching names from a given dataset to a dictionary in which each institution is assigned a unique identifier, and 2) grouping names from a given dataset into multiple clusters, so that each cluster represents a single institution. A common challenge in both approaches is determining whether two names refer to the same institution. Similar names may refer to different institutions, resulting in false-positive matches, whereas dissimilar names may refer to the same institution, resulting in false-negative matches. Therefore, effective disambiguation requires similarity measures that capture institutional identity beyond superficial name similarity.

A popular method for similarity measurement is to identify shared components (e.g., n-grams or characters in names, locations such as cities, states, and countries, as well as components in email addresses) within institution names \citep{ancona2023novel,jonnalagadda2010nemo,shao2020elad}. For instance, Ancona et al. \citep{ancona2023novel} computed the number of common words between name pairs as well as consecutive common characters, which they then used as features for disambiguation. Various auxiliary lexical resources, such as dictionaries and knowledge bases, have also been leveraged to normalize and identify components in institution names. Examples include dictionaries in the Geoworldmap database for country mapping\footnote{\url{http://www.geobytes.com/freeservices.htm}} \citep{jonnalagadda2010nemo} and the Authority File for Affiliations, an affiliation knowledge base containing 113,700 affiliation concepts and approximately 583,700 affiliation names \citep{sun2017using}. The key innovation of these studies typically lies in the rules or algorithms used for similarity measurement between names based on components. Weighting components based on human-crafted rules has been used to measure the similarity of names \citep{morillo2013towards,onodera2011method}. Specifically, the frequency of each component appearing in institution names is calculated, and each component is weighted based on a set of human-crafted rules. The final weighted scores of institution names are used to measure the similarity between names: Institution names with similar weighted scores are considered similar. In addition to simply weighting occurrences of components, algorithms have also been used to compute edit distance, which represents the minimum number of operations required to transform one name into another based on component insertion, deletion, and substitution \citep{huang2014institution}. Normalized compression distance (NCD) has also been introduced, based on the idea that compressing two names together should result in a smaller size than compressing them separately and adding the results \citep{jiang2011affiliation}. A shorter distance indicates higher similarity.

Based on the assumption that an author often publishes multiple papers under the same institution, Huang et al. \citep{huang2014institution} leveraged author information to improve institution name disambiguation. They created an author--institution table where each entry lists an author’s name alongside the corresponding institution names from their publications. They then applied a series of human-crafted rules to this table, based on shared words and edit distances as well as author names, to group institution names into clusters. They tested this multi-method approach, which integrates most of the methods described above, on data for institution names collected from the Web of Science for the domains of mathematics, computer science, psychology, and economics. They achieved precision scores ranging from 0.84 to 0.94 and recall scores from 0.50 to 0.87 across domains, with lower performance in social sciences and higher performance in natural sciences. We speculate that this discrepancy may arise from a broader range of funders in social science research, making funder disambiguation more challenging.

More recently, Dalsgaard et al. \citep{dalsgaard2026linking} developed a large-scale pipeline for linking funder names in WoS to standardized organizations in OpenAlex and ROR. Their approach combines lexical normalization, similarity-based clustering, rule-based matching, named entity recognition, and manual validation. Among 7.4 million unique funder names, 1.9 million were assigned at least one potential match, covering 72\% of all funder mentions in WoS (i.e., a unique funder name may be mentioned repeatedly across publications), while approximately 74\% of unique funder names remained unmatched. These results demonstrate the effectiveness of large-scale linkage for frequently occurring funders while also highlighting the continuing challenge of disambiguating long-tail funder names.

Shared component identification, human-crafted rules, and curated knowledge bases can be effective for certain tasks, but the rules used for institution name disambiguation may not generalize to funder name disambiguation for several reasons. First, institution names are usually followed by well-documented or highly structured informative components, such as locations, zip codes, and domain names from email addresses \citep{guan2025disambiguating}. Second, the level of resolution in funder names can also be more diverse than that in institution names. For example, the name of a research institute typically includes the name of the institution and its sub-component (such as a school or department). In contrast, a funder name might also include the name of a funding program or project. Third, funder names tend to have broader coverage, including more non-research organizations (such as a variety of small businesses and companies). These differences suggest that funder name disambiguation may require more complex rules and extensive knowledge, especially when handling large-scale data. 

With the advancement of large pre-trained models and the growing prominence of fine-tuning techniques in NLP, the potential for using these approaches for funder name disambiguation remains underexplored. Accordingly, we focus on task-specific fine-tuning of pre-trained models for biodiversity conservation publications.

\section{Data}

This study uses three datasets produced by prior efforts to curate funder information.

\begin{itemize}
    \item The Web of Science Core Collection (WoS)\footnote{\url{https://clarivate.com/products/scientific-and-academic-research/research-discovery-and-workflow-solutions/webofscience-platform/web-of-science-core-collection/}} consists of publication records with bibliometric meta-information and funder information. It has recorded funding information since 2008. Funding-related data from 2008 onward are generally considered more complete compared to other data such as Scopus and PubMed \citep{kokol2018discrepancies,liu2020accuracy}. As of February 3, 2022, 122,508 conservation publications were categorized as biodiversity-related papers by the WoS corpus, all of which we downloaded. Biodiversity conservation publications offer three advantages for studying funder name disambiguation, particularly regarding the diversity of funders. First, this domain spans both natural and social sciences as work in this domain is supported by funders from a wide range of fields. Moreover, as noted above, previous research has shown that disambiguating institution names is more challenging in the social sciences than in the natural sciences \citep{huang2014institution}, such that using biodiversity conservation as a domain enhances the generalizability of our work. Second, conservation research involves geographically diverse funding flows. For example, between 2015 and 2022, Africa, Asia, and Latin America and the Caribbean received 30\%, 21\%, and 17\% of biodiversity-related official development finance, respectively \citep{oecd2024biodiversity}. Biodiversity conservation also draws on diverse domestic and international funding sources \citep{cbd2022gbf}, extending the funder landscape beyond major funders in the United States, China, and the European Union. Third, the geographic diversity of funders results in funder names appearing across multiple languages, requiring disambiguation methods to account for multilingual name variations. Specifically, we identified 112 languages in our dataset using fast-langdetect\footnote{\url{https://github.com/LlmKira/fast-langdetect}} \citep{joulin2017bag}. Approximately 85.6\% of the funder names are in English, followed by Portuguese (5.0\%), Spanish (3.8\%), French (1.7\%), and German (1.2\%), while each of the remaining languages accounts for less than 1\% of the funder names. This multilingual distribution introduces additional challenges, including language-specific naming conventions, spelling variations, and transliteration differences. Taken together, the disciplinary diversity, broad global distribution of funders, and multilingual nature of the corpus create a particularly challenging benchmark for funder name disambiguation. A model that performs well on this dataset is therefore expected to generalize more effectively to other heterogeneous, real-world bibliographic data.

    \item The Crossref Open Funder Registry (OFR)\footnote{\url{https://www.crossref.org/services/funder-registry/}} \citep{hendricks2020crossref} is a recent effort to curate funding information. Authors of scholarly publications who use the OFR system can find the unique IDs for the funders they would like to acknowledge, standardize the metadata of their publications, and deposit the standardized metadata in OFR. Compared with WoS, OFR has more standardized funder name records. However, we found that the OFR dataset does not include a significant portion of funder records that are indexed in WoS as mentioned above. In short, WoS provides more comprehensive coverage of funder records than OFR, but its funder names are less standardized and contain more variations: typos (e.g., U.S National Science Foundation), name variants (e.g., National Council for Science and Technology Consejo Nacional de Ciencia y Tecnologia), and different levels of resolution for one funder (e.g., Stanford Medicine, which belongs to Stanford University). Thus, a disambiguation effort is needed for more accurate funding analysis. 

    \begin{figure}[ht]
    \centering
    \includegraphics[width=0.9\textwidth]{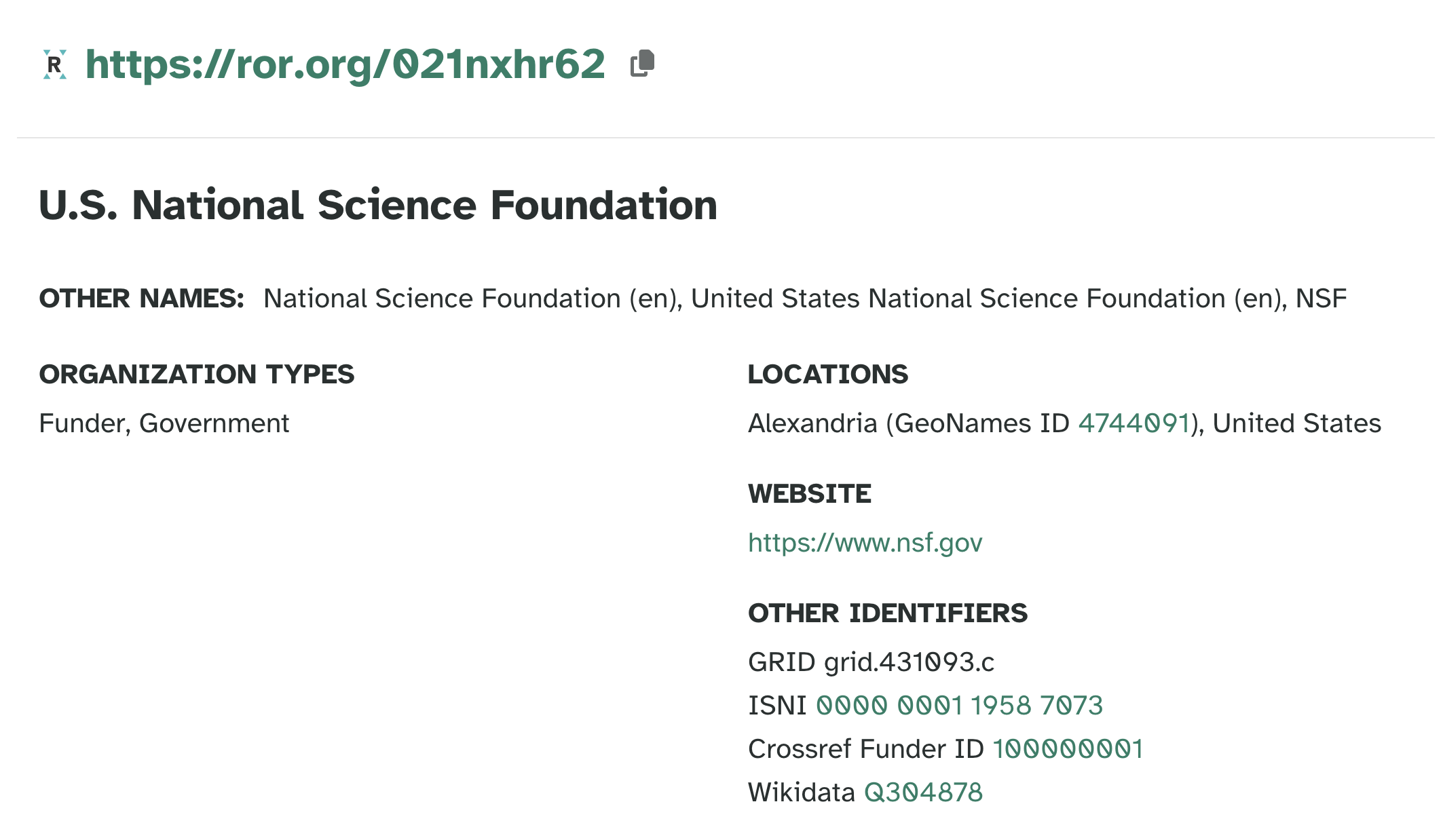}
    \caption{Example of ROR data}
    \label{fig:ror}
    \end{figure} 
    
    \item The Research Organization Registry (ROR)\footnote{\url{https://ror.org/}} \citep{french2022emerging,lammey2020solutions} was launched in 2019 and currently indexes more than 102,000 organizations. Each organization in ROR has a unique ID, a primary name, an associated set of name variants, and meta information such as organization type. Moreover, the inclusion of funder identifiers in other databases (see OTHER IDENTIFIERS in Fig. \ref{fig:ror}) helps to establish connections between them. This dataset is a dictionary for organizational information and does not contain bibliometric metadata.
\end{itemize}

\section{Methodology}

\subsection{Detailed Workflow}

To obtain cleaner and more comprehensive funding information for analysis, we use two complementary approaches. For funders indexed in well-curated organization registries with unique identifiers (e.g., ROR), we map WoS funder names to their corresponding records based on name similarity. For funders not indexed in ROR, we cluster funder names based on their similarity, with each cluster representing a distinct funder. One strategy for mapping and clustering funder names is to directly use Sentence Transformer (ST) models, which are a widely used for semantic search due to its stable performance for tasks based on textual similarity calculation \citep{reimers-2019-sentence-bert}. ST models generate text embeddings that capture semantic similarity between texts. The original ST model was fine-tuned on Natural Language Inference (NLI) data by combining the SNLI \citep{bowman2015large} and MultiNLI \citep{williams2018broad} datasets, totaling approximately one million manually labeled sentence pairs (e.g., ``A smiling costumed woman is holding an umbrella'' vs. ``A happy woman in a fairy costume holds an umbrella''). A variety of ST models have been subsequently trained or fine-tuned on broader and more diverse datasets, commonly with pairs of texts that capture different types of semantic relationships. These models compute the cosine similarity between the embeddings of two candidate strings. Other embedding models with similar functionality but different architectures have also emerged in recent years, including those built upon the widely adopted Qwen3 \citep{zhang2025qwen3} and Gemma \citep{vera2025embeddinggemma} families. Although these foundation models demonstrate strong general-purpose matching capabilities, they are not tailored to the domain-specific nuances of funder name disambiguation, necessitating fine-tuning on our target task. Given that our dataset encompasses funder names across multiple languages, we selected paraphrase-multilingual-mpnet-base-v2\footnote{\url{https://huggingface.co/sentence-transformers/paraphrase-multilingual-mpnet-base-v2}} (a multilingual ST model), EmbeddingGemma\footnote{\url{https://huggingface.co/google/embeddinggemma-300m}}, and Qwen3-Embedding\footnote{\url{https://huggingface.co/Qwen/Qwen3-Embedding-0.6B}} models as our starting models.

\begin{figure}[!htbp]
    \centering
    \includegraphics[width=1\textwidth]{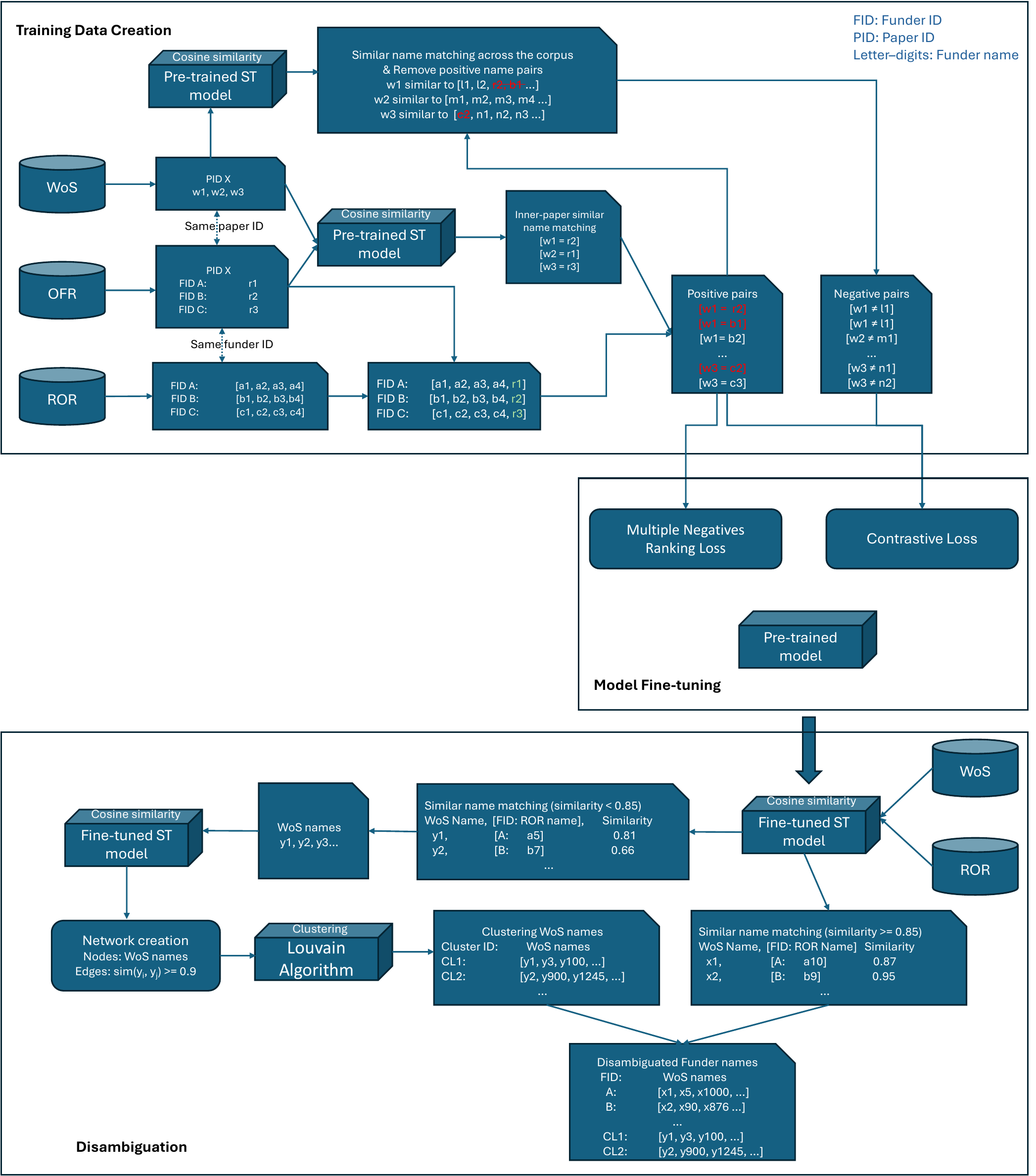}
    \caption{Steps for training data creation, model fine-tuning, and funder name disambiguation. \textbf{Paper} refers to a published research article. \textbf{Funder name} refers to a single textual representation of a funding organization as it appears in a paper. \textbf{Funder ID} refers to the unique identifier assigned to a funder. Multiple funder names across different papers may refer to a single funder. Therefore, we use the funder ID to link these name variants to the same underlying entity.}
    \label{fig:dis_workflow}
\end{figure}

Figure \ref{fig:dis_workflow} illustrates the detailed workflow of the disambiguation process, which consists of three main parts: training data creation, model fine-tuning, and application of the fine-tuned model for funder name disambiguation.

\subsection{Training Data Restructuring}
A major challenge in building a funder name disambiguation model is the lack of training data with funder information for model training or fine-tuning. Manual curation of a large-scale training set is time-consuming and costly, because WoS contains a very large number of funders and name variants. We thus created a training set by restructuring existing data, namely WoS, OFR, and ROR. In the training set, each instance represents a pair consisting of an anchor name (input name) and a positive/negative name as well as a label indicating whether or not these two names refer to the same funder (positive or negative pair): [name 1, name 2, 1 (positive) or 0 (negative)].

To create positive pairs, we used OFR as a bridge between WoS and ROR. ROR and OFR can be linked directly because both sources include Crossref Funder IDs for individual funder names in their records. However, WoS and OFR can be linked only at the paper level through Digital Object Identifiers (DOIs), not at the level of individual funders. Funder records for the same paper may still differ across the two datasets. For example, a paper in the OFR data might list ``University of Illinois at Urbana-Champaign'' and ``National Science Foundation'' as funders, while the same paper in the WoS data might list ``the Graduate College of the University of Illinois at Urbana-Champaign'' and ``U.S. National Science Foundation'' as funders. Fortunately, matching names within a single paper is substantially easier than matching them across papers, primarily because each paper contains relatively few funders, with a median of two. The pre-trained ST model can effectively match funder names within a single paper. Within-paper matches with cosine similarity above 0.70 achieved accuracy above 0.98, we retained all such pairs. This process assigned a unique ID to funder names in 20,793 of the 52,760 papers. We then obtained all possible positive pairs among names sharing the same ID, yielding 54,517 positive pairs.

After identifying the positive pairs, we treated the remaining funder name combinations, other than positive pairs, as negative pairs in the 20,793 papers. For each funder name, we retrieved its 20 most similar names and removed any known positive pairs. The remaining high-similarity pairs served as hard negatives. We excluded lower-similarity negative pairs, or soft negatives, because prior work has found that hard negatives provide stronger learning signals \citep{yang2024does}. This procedure yielded 215,163 negative pairs.

Given that the positive and negative pairs we created might be limited in range and scope as they are from 20,793 out of 52,760 papers, we applied the same methods solely to ROR to expand the funder list and identified 140,684 positive pairs and 3,061,200 negative pairs. Of these ROR-derived pairs, we used all positive pairs and 500,000 randomly selected negative pairs. This is because we found that the pre-trained model performs well (98\% accuracy) for funder names in ROR, as these names are cleaner than those in WoS. Including substantially more such pairs would increase the training cost while providing little additional improvement.

Finally, we created and selected 195,201 (54,517 from WoS + 140,684 from ROR) positive pairs and 715,163 (215,163 from WoS + 500,000 from ROR) negative pairs. Among these, 5,000 positive pairs were randomly sampled for out-of-sample testing, and the remaining 905,364 pairs were used for model fine-tuning (40,000 for validation and 865,364 for training).

\subsection{Multi-task Learning}
Instead of using a single loss function for fine-tuning, we adopted a multi-task learning approach that involves two loss functions for two key reasons. First, our training set contains two types of funder name pairs (positive and negative), with an imbalance of approximately 3.7 negative pairs for every positive pair. Employing one loss function to learn primarily from positive pairs and another from both positive and negative pairs allows us to better manage this imbalance. Second, given that WoS includes many small or lesser-known funders that cannot be mapped to ROR, our model must both map funder names to ROR and determine whether a valid ROR match exists. Using one loss function to update the parameters for mapping and another to update those for classification enables the model to handle both tasks effectively. Therefore, we utilized both Contrastive Loss (CL) and Multiple Negatives Ranking Loss (MNRL).

\begin{itemize}
    \item Contrastive Loss \citep{hadsell2006dimensionality}: This loss function uses both positive and negative pairs and can be written as:

    \begin{equation}
    \mathcal{L}_{\text{CL}} = \frac{1}{2} \left[ y_i \cdot D(u_i, v_i)^2 + (1 - y_i) \cdot \max\bigl(0, m - D(u_i, v_i)\bigr)^2 \right]
    \end{equation}

    In the equation, $D(u_i, v_i) = 1 - S(u_i, v_i)$ represents the cosine distance derived from the cosine similarity $S(u_i, v_i)$ between the two normalized embeddings ($u_i$ and $v_i$). The binary label $y_i \in \{0, 1\}$ indicates whether the two names refer to the same funder entity ($y_i = 1$) or distinct entities ($y_i = 0$). The margin $m$ acts as a minimum separation threshold, penalizing dissimilar pairs whenever their distance falls below $m$ ($m = 0.5$ in our implementation, corresponding to an upper cosine similarity threshold of $1 - m = 0.5$). This loss function is particularly effective at separating negative funder pairs in the embedding space\footnote{Technically, contrastive loss reduces the distance between matching funder names as well. In practice, however, its primary strength lies in separating dissimilar pairs, actively penalizing hard negative pairs that fall within the margin radius rather than merely pulling positive pairs closer together.} by ensuring their distance remains above the required distinguishing threshold.

    \item Multiple Negatives Ranking Loss \citep{henderson2017efficient}: It exclusively uses positive pairs in our training set as input. In each batch, there are $K$ positive pairs, ranging from ($u_1$, $v_1$) to ($u_k$, $v_k$), where $u_i$ is the $i^{th}$ anchor name, and $v_i$ is the $i^{th}$ positive name. MNRL assumes that $u_i = v_i$ and $u_i \neq v_j$ when $i \neq j$. Therefore, for each anchor name $u_i$, there is one positive name ($v_i$) and K-1 negative names ($v_j$). For a single batch, it can be expressed as:

    \begin{equation}
    \mathcal{L}_{\text{MNRL}} = -\frac{1}{K}\sum_{i=1}^K \log P_{\text{approx}}(v_i \mid u_i) = -\frac{1}{K}\sum_{i=1}^K \left[ S(u_i, v_i) - \log \sum_{j=1}^K e^{S(u_i, v_j)} \right]
    \end{equation}

    In the equation, $S(u_i, v_i)$ represents the cosine similarity between the two embeddings ($u_i$ and $v_i$). $\sum^K_{j=1}e^{S(u_i,v_j)}$ is the sum of the exponentiated cosine similarity scores for all $K$ embeddings. It is the key part of the softmax normalization, where it normalizes raw scores into a probability distribution. Therefore, this loss function aims to minimize the approximate mean negative log-likelihood of the data. This loss function is particularly effective at minimizing the distance between positive names because its task is to identify the positive name of each anchor name from a list of $K$ candidates (220 in our case\footnote{Technically, a larger $K$ could lead to higher matching accuracy. However, it will require larger GPU memory.}).
\end{itemize}

\subsection{Baselines}
\begin{itemize}
    \item \textbf{Pre-trained embedding models:}
    Widely-used pre-trained embedding models are evaluated as embedding-based retrieval baselines. Given a funder name, each model generates a dense embedding that is compared against candidate funder names using cosine similarity. The evaluation is restricted to models specifically optimized for semantic matching tasks because our preliminary experiments with general-purpose encoder models such as BERT and RoBERTa achieved an accuracy below 0.3.

    The evaluated models include both traditional sentence embedding models and recent large language model (LLM)-based embedding models:

    \begin{itemize}
        \item \textbf{ST Models}: An English model, \textit{all-mpnet-base-v2}\footnote{\url{https://huggingface.co/sentence-transformers/all-mpnet-base-v2}}, and a multilingual model, \textit{paraphrase-multilingual-mpnet-base-v2}, which supports more than 50 languages.
        
        \item \textbf{EmbeddingGemma}: A lightweight embedding model developed by Google DeepMind, recognized as one of the most capable models under 0.5 billion parameters \citep{vera2025embeddinggemma}. For brevity, we refer to this model as \textbf{Gemma} throughout the remainder of this paper.

        \item \textbf{Qwen3-Embedding}: A family of recent embedding models based on the Qwen3 architecture \citep{zhang2025qwen3}. We evaluated the 0.6B, 4B, and 8B variants to investigate how model size influences matching performance. Throughout this paper, \textbf{Qwen3} refers specifically to these embedding models rather than the generative LLMs.
    \end{itemize}

    \item \textbf{Generative LLMs:} We also evaluated state-of-the-art generative LLMs (GPT-5.2, Gemini-2.5-Flash, and Claude-Sonnet-4.6). We used the following prompt to identify the official name for each funder in our test set:
    
        \textit{Please let me know the official name of the following funder or organization. Only return one name without any acronym. Funder name: \textless Name from WoS\textgreater \ Official name: }

        Since generative LLMs may return different official names for the same funder (e.g., ``Alexander von Humboldt-Stiftung'' and ``Alexander von Humboldt Foundation'' or ``Natural Sciences and Engineering Research Council of Canada'' and ``Natural Sciences and Engineering Research Council''), the generated names still require post-processing to map them to entities in ROR. We experimented with two post-processing methods: 
    
    \begin{itemize}
        \item \textbf{Normalized exact matching}, which normalizes both the generated name and ROR names by lowercasing and removing special characters before performing exact string matching.
        
        \item \textbf{Embedding-based matching}, which combines a generative LLM with a pre-trained multilingual embedding model to encode the generated official name and retrieve the most semantically similar ROR entity based on cosine similarity.
    \end{itemize}
\end{itemize}

\section{Matching Accuracy on Test Data}
Using the fine-tuned models, the cosine similarity is computed between each anchor funder name in the test data and every funder name in the ROR corpus. The funder name with the highest cosine similarity score in the ROR corpus is considered a match for the anchor name. Since each matched name in the ROR corpus has a unique funder ID, the anchor name inherits this ID. Finally, the anchor funder name is considered correctly disambiguated if its inherited ID matches the ID of the corresponding positive name in the test data.

\begin{table}[h!]
\centering
\begin{tabular}{l|cc|cc}
\toprule
& \multicolumn{2}{c|}{\textbf{ROR}} & \multicolumn{2}{c}{\textbf{WoS}} \\
\textbf{Model} & \textbf{All} & \textbf{Unique} & \textbf{All} & \textbf{Unique} \\
\midrule

\multicolumn{5}{c}{\textbf{Pre-trained Embedding Model}} \\
\cmidrule(lr){1-5}
$\text{ST}_{Eng}$ & 0.97 & 0.97 & 0.47 & 0.33 \\
$\text{ST}_{multi}$ & \textbf{0.98} & \textbf{0.98} & 0.70 & 0.66 \\
Gemma & \textbf{0.98} & \textbf{0.98} & \underline{0.76} & \underline{0.74} \\
Qwen3-0.6B & \textbf{0.98} & \textbf{0.98} & 0.69 & 0.68 \\
Qwen3-4B & \textbf{0.98} & \textbf{0.98} & 0.65 & 0.63 \\
Qwen3-8B & \textbf{0.98} & \textbf{0.98} & 0.68 & 0.68 \\

\midrule
\multicolumn{5}{c}{\textbf{Generative LLM + Exact Matching}} \\
\cmidrule(lr){1-5}
$\text{GPT-5.2}$ & \underline{0.70} & 0.69 & 0.63 & 0.55 \\
$\text{Gemini-2.5-Flash}$ & \underline{0.70} & \underline{0.70} & \underline{0.67} & 0.59 \\
$\text{Claude-Sonnet-4.6}$ & \underline{0.70} & 0.60 & 0.60 & \underline{0.60} \\

\midrule
\multicolumn{5}{c}{\textbf{Generative LLM + Pre-trained Embedding Model}} \\
\cmidrule(lr){1-5}
$\text{GPT-5.2 + Gemma}$ & \underline{0.93} & \underline{0.93} & 0.81 & \underline{0.75} \\
$\text{Gemini-2.5-Flash + Gemma}$ & 0.87 & 0.87 & \underline{0.82} & \underline{0.75} \\
$\text{Claude-Sonnet-4.6 + Gemma}$ & 0.84 & 0.84 & 0.81 & 0.74 \\

\midrule
\multicolumn{5}{c}{\textbf{Fine-Tuned Embedding Model}} \\
\cmidrule(lr){1-5}
$\text{ST}$ & \textbf{0.98} & \textbf{0.98} & \textbf{0.91} & \textbf{0.89} \\
Gemma & \textbf{0.98} & \textbf{0.98} & \textbf{0.91} & \textbf{0.89} \\
Qwen3-0.6B & \textbf{0.98} & \textbf{0.98} & 0.85 & 0.77 \\

\midrule
\multicolumn{5}{c}{\textbf{GPT-5.2 + Fine-tuned Embedding Model}} \\
\cmidrule(lr){1-5}
GPT-5.2 + ST & \underline{0.92} & \underline{0.92} & \underline{0.86} & \underline{0.82} \\
GPT-5.2 + Gemma & 0.91 & 0.91 & 0.84 & 0.79 \\
GPT-5.2 + Qwen3-0.6B & 0.91 & 0.91 & 0.83 & 0.75 \\

\bottomrule
\end{tabular}
\caption{Model performance in terms of accuracy. Bold values denote the highest accuracy in each column across all approaches, and underlined values denote the highest accuracy within each approach.}
\label{tab:model_performance}
\end{table}

Table \ref{tab:model_performance} presents the matching accuracy on the test data. The test dataset is divided into two subsets according to whether the anchor funder name originated from ROR or WoS. The ROR names are generally clean and standardized because they are curated organization names, whereas the WoS names are substantially noisier, as many are extracted directly from funding acknowledgment texts. Additionally, because a funder may appear multiple times in the test data under different name variants, we constructed an alternative dataset (denoted as ``Unique'') by retaining only one occurrence of each funder.

\textbf{Frequently occurring funders are easier to disambiguate.}
Across nearly all models, accuracy is consistently higher on the complete WoS dataset (denoted as ``All'') than on the unique subset, where duplicate funders have been removed. This observation suggests that frequently occurring funder names are easier to identify than rare funders. One possible explanation is that common funders contribute more training examples during fine-tuning. In addition, for pre-trained models, these funders might be more likely to appear in the large-scale corpora used to pre-train modern language models, enabling the models to learn richer semantic representations for well-known funders than for rarely occurring funders.

\textbf{Pre-trained embedding models perform well on standardized funder names but struggle with noisy funder names.}
For the ROR data, all embedding models, including the Multilingual ST model, Gemma, and the Qwen3 family, achieve an accuracy of approximately 0.98, comparable to our fine-tuned models. This result suggests that recent embedding models are already highly effective for matching clean, standardized funder names curated in ROR. However, their performance drops substantially on the WoS data. For example, the best-performing pre-trained model (Gemma) achieves only an accuracy of 0.76 on the WoS data. These results indicate that the primary challenge in funder name disambiguation lies in handling noisy, real-world funder names rather than standardized ones.

\textbf{Generative LLMs are not suitable as standalone models for funder name disambiguation.}
Using generative LLMs followed by exact matching yields substantially lower accuracy than embedding-based approaches, achieving only around 0.7 on the ROR data and 0.6--0.67 on the WoS data. Unlike embedding models specifically designed for semantic retrieval and similarity matching, generative LLMs do not consistently produce the canonical funder names required for reliable exact matching against the ROR data. These results suggest that generative LLMs alone are not well suited for large-scale funder name disambiguation.

\textbf{Generative LLMs can improve the standardization of noisy funder names, but may also propagate errors to downstream matching.}
Combining generative LLMs with pre-trained embedding models consistently improves performance on the WoS data, increasing the accuracy from 0.76 to 0.81--0.82. This improvement suggests that generative LLMs can effectively standardize noisy funder names before semantic matching. However, the same approach performs substantially worse on the standardized ROR data, where the best pre-trained embedding models achieve approximately an accuracy of 0.98, compared with only 0.84--0.93 for the hybrid approaches. The same pattern is observed when generative LLMs are combined with fine-tuned embedding models. These results suggest that generative LLMs can only improve funder name matching when they can successfully standardize noisy funder names. However, when the standardization is incorrect, the error is inevitably propagated to the downstream embedding matching stage. Therefore, generative LLMs could reduce the overall matching accuracy. In other words, generative LLMs represent a double-edged sword for funder name disambiguation: They can improve performance by standardizing noisy funder names, but LLM errors can also propagate to the matching step and degrade the final results.

\textbf{Our fine-tuning approach substantially improves the funder name matching task.}
On the WoS data, the fine-tuned ST and Gemma models achieve an accuracy of approximately 0.91 for all funder names and 0.89 for unique funder names, outperforming their corresponding pre-trained models by 0.15--0.21 and 0.15--0.23, respectively. The fine-tuned Qwen3-0.6B model also substantially improves the accuracy, although its performance is lower than the other two fine-tuned models. These consistent improvements across multiple embedding architectures demonstrate that our fine-tuning approach generalizes well and is not limited to a particular model. The fine-tuned ST and Gemma models also outperform the best hybrid approach (generative LLM + pre-trained embedding model) by 0.09 for all funder names and 0.14 for unique funder names, respectively. Overall, these results demonstrate that domain-specific fine-tuning is an effective strategy for funder name matching.

\textbf{Larger embedding models do not necessarily achieve better performance.}

Larger pre-trained models do not consistently outperform smaller ones. Before fine-tuning, Gemma achieves better performance than the larger Qwen3 models on the WoS dataset. Even within the Qwen3 family, increasing model size does not consistently improve performance. After fine-tuning, the ST and Gemma models also outperform the fine-tuned Qwen3-0.6B model despite their smaller sizes. These results suggest that simply increasing model size does not necessarily lead to better funder name matching performance. Possible explanations include differences in model architectures and pre-training objectives, as well as the possibility that larger models require more training data or different fine-tuning strategies to fully realize their potential \citep{liu2025not,zhang2024scaling}.

\section{Disambiguation on the WoS Corpus}
\subsection{Distinguishing Funder Names within ROR's Coverage}

We applied the fine-tuned ST model to the WoS corpus because of its best performance, retrieving the most similar funder name and its corresponding ID from the ROR data. Unlike the test data, the WoS corpus includes funder names that extend beyond ROR's coverage. This means that in addition to matching funder names, we must also identify those that are outside the ROR's coverage, and thus cannot be disambiguated by mapping them to the ROR data.

\begin{table}[h!]
\centering
\begin{tabular}{|c|c|c|c|c|}
\hline
\textbf{Cosine similarity} & \textbf{$<$0.8} & \textbf{0.8-0.85} & \textbf{0.85-0.9} & \textbf{$\geq$ 0.9} \\ \hline
\textbf{Matching accuracy} & 0.0625       & 0.38            & 0.75            & 0.98               \\ \hline
\textbf{Instance proportion}          & 0.25       & 0.125           & 0.105           & 0.52               \\ \hline
\end{tabular}
\caption{Cosine similarity, matching accuracy, and proportion of instances}
\label{tab:cosine_similarity}
\end{table}

Given that our model is partially fine-tuned using Contrastive Loss, we can establish a threshold similarity score, below which a WoS funder name is unlikely to be matched with any names in the ROR data. We categorized the data into four groups based on the cosine similarity score between each funder name in WoS and its matched name in ROR: $<$0.8, 0.8-0.85, 0.85-0.9, and $\geq$ 0.9. We then randomly selected around 200 instances from each group and manually annotated whether the matching result was correct. As shown in Table \ref{tab:cosine_similarity}, 52\% of the WoS funder names have a match in ROR with a similarity score of $\geq$ 0.9, achieving a matching accuracy of 0.98. The accuracy drops to 0.75, 0.38, and 0.0625 for similarity scores of 0.85-0.9, 0.8-0.85, and $<$ 0.8, respectively.

We evaluated cosine similarity scores of 0.80, 0.85, and 0.90 as classification thresholds. For each threshold, funder names with similarity scores below the threshold were classified as unmatched, whereas those with scores equal to or above the threshold were classified as matched. For illustration, the equations below show the calculation for a threshold of 0.85. We applied the same procedure to thresholds of 0.80 and 0.90 by regrouping the similarity intervals according to whether they fell below or at or above each threshold.

\begin{equation}
True~positives = (M_{\geq0.9} \cdot I_{\geq0.9} +  M_{0.85-0.9} \cdot I_{0.85-0.9})  \cdot n
\end{equation}
\begin{equation}
False~positives = ((1-M_{\geq0.9}) \cdot I_{\geq0.9} +  (1-M_{0.85-0.9}) \cdot I_{0.85-0.9}) \cdot n
\end{equation}
\begin{equation}
True~negatives = ((1-M_{0.8-0.85})\cdot I_{0.8-0.85} + (1-M_{<0.8})\cdot I_{<0.8}) \cdot n
\end{equation}
\begin{equation}
False~negatives =  ((M_{0.8-0.85} \cdot I_{0.8-0.85} + M_{<0.8} \cdot I_{<0.8})) \cdot n
\end{equation}

where $M_i$ and $I_i$ denote the correct match rate and the proportion of instances for group $i$, and $n$ is the number of instances in the corpus.

We then calculated the precision, recall, and F1 scores for our classification strategy using different thresholds. As shown in Table \ref{tab:threshold_metrics}, a cosine similarity threshold of 0.85 yields the highest F1 score (0.92) and provides the best balance between precision (0.94) and recall (0.90). Applying this threshold yielded a disambiguated dataset covering 93,155 (62.6\% of) funder names in the WoS corpus with an F1 score of 0.92. These names correspond to 7,953 unique funders indexed in ROR. The remaining set contained 55,635 funder names, representing 37.4\% of the funder corpus, including an estimated 9,392 false negatives that should have been matched to ROR and 46,243 true negatives that were not indexed. In the following sections, we will refer to the disambiguated dataset (62.6\%) as the main dataset and the remaining one as the remaining dataset (37.4\%).

\begin{table}[h!]
\centering
\begin{tabular}{|c|c|c|c|}
\hline
\textbf{Acceptance threshold} & \textbf{Precision} & \textbf{Recall} & \textbf{F1 score} \\ \hline
$\geq 0.90$ & 0.980 & 0.782 & 0.870 \\ \hline
$\geq 0.85$ & 0.941 & 0.903 & 0.922 \\ \hline
$\geq 0.80$ & 0.848 & 0.976 & 0.907 \\ \hline
\end{tabular}
\caption{Estimated precision, recall, and F1 scores under different cosine similarity acceptance thresholds.}
\label{tab:threshold_metrics}
\end{table}

\subsection{Clustering Funder Names Beyond ROR's Coverage}
Although the previous steps could not confidently match WoS funder names in the remaining dataset to the ROR data, we can still cluster the unmatched funder names using our model to group funder names for a second round of disambiguation and also to understand why the prior disambiguation failed (i.e., what types of funders are not indexed by ROR, and why mismatches and misclassifications occurred). We computed the cosine similarity between every pair of names in the remaining dataset. If the similarity score between two funder names was 0.9 or higher, we established a link between them, ultimately forming a funder name network. In this network, funder names are represented as nodes, links with a similarity score of 0.9 or higher as edges, and similarity scores as edge weights. We then applied the Louvain algorithm \citep{blondel2008fast} for network community detection, identifying 28,357 unique funders. Each community or cluster was considered to represent either a single funder or a group of similar funders.

The occurrences of funders in prior conservation publications are highly right-skewed. Only 1.5\%, 2.9\%, and 14.1\% of funders in the main (matching) dataset appear in at least 100, 50, and 10 publications, respectively, while 46.8\% of funders occur in only one publication. For the remaining (clustering) dataset, according to our clustering results, these percentages are 0.067\%, 0.19\%, 1.66\%, and 76.2\%, respectively. The more right-skewed distribution in the remaining dataset compared to the main dataset indicates that a greater number of funders in the remaining dataset occur infrequently. Examples of these funders that appear once in our data include ``Instituto Biotropicos'' (a small NGO from Minas Gerais, Brazil), ``Pride of Maui'' (a small travel company from Wailuku, Hawaii), and ``Guizhou R\&D Program for Social Development'' (a local government fund from Guizhou, China). The high proportion of infrequently occurring funders reflects the long-tail nature of the funding landscape, where a diverse range of small, local, and specialized organizations contribute to research funding but may have limited representation in existing crowdsourced resources such as ROR, making their names more challenging to disambiguate.

\begin{table}[h!]
\centering
\small
\setlength{\tabcolsep}{5pt}
\renewcommand{\arraystretch}{1.2} 
\begin{tabularx}{\textwidth}{>{\raggedright\arraybackslash}X r c | >{\raggedright\arraybackslash}X r c}
\toprule
\multicolumn{3}{c|}{\textbf{Main Data -- Matching}} & \multicolumn{3}{c}{\textbf{Remaining Data -- Clustering}} \\
\cmidrule(r){1-3} \cmidrule(l){4-6}
\textbf{Funder} & \textbf{Count} & \textbf{Region} & \textbf{Funder} & \textbf{Count} & \textbf{Region} \\
\midrule
National Science Foundation & 4,560 & U.S. & European Framework Programme & 544 & EU \\
\hline
National Natural Science Foundation of China & 3,402 & China & European Regional Development Fund & 426 & EU \\
\hline
National Council for Scientific and Technological Development & 2,804 & Brazil & Ministry of Science and Innovation & 386 & Spain \\
\hline
Natural Environment Research Council & 2,674 & U.K. & National Basic Research Program of China & 370 & China \\
\hline
Natural Sciences and Engineering Research Council of Canada & 2,068 & Canada & Darwin Initiative & 257 & U.K. \\
\hline
Brazilian Federal Agency for Support and Evaluation of Graduate Education & 1,666 & Brazil & DST-NRF Centre of Excellence for Invasion Biology -- India & 227 & India \\
\hline
U.S. Fish and Wildlife Service & 1,565 & U.S. & National Key R\&D Program of China & 203 & China \\
\hline
European Commission & 1,379 & EU & Ministry of Education and Science & 189 & Spain \\
\hline
Foundation for Science and Technology & 1,192 & Portugal & Pittman--Robertson Federal Aid in Wildlife Restoration Act & 180 & U.S. \\
\hline
Australian Research Council & 1,143 & Australia & Program for New Century Excellent Talents in University & 171 & China \\
\bottomrule
\end{tabularx}
\caption{Top 10 most frequently occurring funders in the main and remaining datasets, identified through matching and clustering, respectively.}
\label{tab:frequent_funders}
\end{table}

Table \ref{tab:frequent_funders} presents the most frequently occurring funders identified through matching in the main dataset and clustering in the remaining dataset, respectively. Seven of the ten most frequently occurring funders in the clustering dataset are outside the scope of ROR. For instance, we identified various funders affiliated with the European Commission/Union in the first two clusters, including two major funders—``European Framework Programme'' and ``European Regional Development Fund''—neither of which was indexed by ROR. Similarly, some major Chinese funders are also missing from ROR, such as the ``National Basic Research Program of China'', ``National Key R\&D Program of China'', and the ``Program for New Century Excellent Talents in University.'' Whether a ``program'' should be regarded as an organization is debatable, but in some countries or regions like EU and China, researchers frequently report large funding programs as funders, further complicating funder name disambiguation. This also contributes to their absence from ROR, as ROR typically excludes programs that fall outside its organizational scope.

Among the three funders that are within the scope of ROR, only one, ``Darwin Initiative'', is directly indexed by ROR. Our fine-tuned model matched this funder to ROR in most instances; however, the similarity score was only around 0.8 (below the threshold of 0.85), leading to the omission of these correct matches. This lower similarity score is primarily due to variations in the funder's name, such as ``Darwin Initiative for the Survival of Species'', which includes additional long postfixes that reduce the similarity score.
Two cases, ``DST-NRF Centre of Excellence for Invasion Biology (India)'' and ``Spanish Ministry of Science and Innovation'', require further discussion. ``DST-NRF Centre of Excellence for Invasion Biology'' typically refers to a funder in South Africa, yet the WoS dataset frequently associates this name with an organization in India. It is unclear whether these refer to the same funder, as there is little information about the Indian organization available online. ``Spanish Ministry of Science and Innovation'' is frequently mentioned in the WoS corpus, yet its official name should be ``Ministry of Science, Innovation and Universities''. However, this name is often matched to ``Ministry of Science and Innovation'' in other countries by our model with a low similarity score.

These examples demonstrate that clustering helps identify major funders not indexed by ROR and correct some mismatches. 
Furthermore, a comparison of the most frequently occurring funders in the well-disambiguated main dataset versus the remaining dataset reveals that funders from non-English-speaking countries present additional challenges. Many of these funders are not indexed in ROR. Additionally, funder names from non-English-speaking countries are often written in other languages (though English names are provided in the table), and the definition of ``funder'' may vary, with some regions using terms like ``programs'' to refer to funders.

\section{Discussion and Conclusion}

In this paper, we introduced (1) a comprehensive and reusable framework for training-data creation, model fine-tuning, and funder-name disambiguation across multiple model architectures, including Sentence Transformer, Qwen3, and Gemma models; (2) empirical evidence that fine-tuned embedding models substantially outperform pre-trained embedding models and generative LLMs on this task; (3) a large-scale dataset of biodiversity conservation publications with cleaned and standardized funder names; and (4) an analysis of the opportunities and challenges in funder-name disambiguation, including limitations in existing funder records and organizational information.

\textbf{The added value of combining multiple funder-related data sources:} Previous studies have typically concentrated on a single source of publication records, occasionally supplemented with dictionaries, ontologies, or registries related to organizational information to support rule-based disambiguation \citep{ancona2023novel,jonnalagadda2010nemo,shao2020elad}. Our research demonstrates that linking and matching multiple sources of publication records (such as WoS and OFR) with data or tools related to organizational information (such as ROR) can facilitate the creation of training datasets for model fine-tuning or training. The lack of supervised learning efforts in previous organization name disambiguation work could be attributed to the absence of large-scale training data. Our approach introduces a method for training data creation without requiring costly manual annotation.

\textbf{Multi-task learning and multi-functional model:} We experimented with multi-task learning for model fine-tuning, resulting in a multi-functional model capable of handling several steps in funder name disambiguation. These functions include mapping funder names in WoS to those in ROR, classifying whether funder names are mappable in the first phase, and clustering WoS funder names when they are unlikely to be indexed in ROR. Since no single step can resolve all disambiguation challenges, training or fine-tuning a multi-functional model can be highly beneficial.

\textbf{Challenge of disambiguating rarely occurring funders and those from non-English-speaking countries:} The greatest challenge is to disambiguate funders that occur infrequently, particularly small, local, or specialized organizations. Pre-trained large models may not capture sufficient information about these funders, even with large-scale training data. Also, creating our own training data that includes funder information is challenging because these funders appear too infrequently in publication records and are not indexed by curated data related to organization or funder information. One possible approach might be to label them as unknown or uncertain when analyzing them and use statistical methods to mitigate their negative impact on analysis. Funders from non-English-speaking countries present a related challenge. Although some of these funders may occur frequently in scholarly publication data, they may still have limited representation in existing curated resources like ROR because of differences in language, naming conventions, and funding systems. For example, some regions frequently use the term ``program'' to refer to funding entities, which can complicate their representation and disambiguation in organization-focused resources. Our work suggests that clustering can help identify some frequently occurring funders that are not represented in curated resources.

\textbf{Limitations and future work:} This work has a significant limitation, in addition to the challenges previously discussed: Some correct matches (e.g., ``Darwin Initiative'' vs. ``Darwin Initiative for the Survival of Species'') identified in the first step of the matching task may be filtered out in the second step of the classification task due to low similarity scores caused by lengthy prefixes or suffixes. Generative LLMs could potentially assist in splitting lengthy names and extracting the core components before applying our model for matching. In other words, while our experiments demonstrate that generative LLMs do not perform well when used directly for name disambiguation (i.e., official name prediction), they may still be useful for supporting intermediate steps in the process, such as name cleaning and cluster verification. Future work can explore integrating generative LLMs into these intermediate stages to improve the robustness and accuracy of the overall disambiguation process.

\textbf{Cautious Use of the Model and Data:} Given the challenges and limitations discussed above, it is important to acknowledge that disambiguation models are not infallible and will still make errors. We highly recommend that users of the model and data follow these steps before conducting funder analysis:

\begin{itemize}
    \item The ROR dataset contains over 102,000 organization names across various domains and countries, some of which may not be relevant to specific studies. Users should consider removing organizations that are not of interest before using the model for matching. Narrowing down the pool of matching candidates can improve matching accuracy. Additionally, we strongly recommend that users remove acronyms from the ROR data when matching WoS names to ROR names, as many funders share the same acronym. For example, NSF can represent both the National Science Foundation and the National Sleep Foundation.

    \item We set a threshold of 0.85 for the similarity score to classify whether a funder is accurately matched to ROR, aiming for a high F1 score. This allowed us to split the data into the main dataset, which is accurately matched, and the remaining dataset, which likely contains funders not indexed by ROR. Since a higher threshold increases precision but decreases recall (and vice versa), users should adjust the threshold based on their research priorities—whether they prioritize precision, recall, or F1 score.

    \item When constructing similarity networks for clustering funder names in the remaining dataset, we used a threshold of 0.9. Users can also modify this threshold according to their needs. A higher threshold will result in smaller, more specific clusters, while a lower threshold will produce larger clusters.

    \item Finally, we strongly recommend that users review the data or the model results and make necessary manual adjustments based on their needs before conducting any funder analysis.

\end{itemize}

\section{Data and Code} \label{data}
The datasets utilized in this study are derived from the Web of Science (WoS) and are therefore subject to Clarivate’s data licensing agreements. These terms prohibit the direct redistribution of raw publication records alongside our disambiguated funder names. To maintain compliance while supporting reproducibility, we provide a mapping file containing only publicly accessible identifiers: article titles, WoS IDs and DOIs. This allows researchers with authorized access to WoS to reconstruct the complete dataset. Furthermore, we have made our implementation code available to the community. All source code and the aforementioned identifier data can be accessed in our GitHub repository\footnote{\url{https://github.com/khan1792/Funder_name_disambiguation}}. Researchers with appropriate access to the underlying WoS data may contact the authors to inquire about access to the disambiguated dataset and fine-tuned models.

\section{Funding Statement}

This research was supported by the John D. and Catherine T. MacArthur Foundation (Grant No. 18-1802-152800-CSD).

\bibliographystyle{SageV} 
\bibliography{reference}

@inproceedings{ali2022named,
  title={Named entity recognition using deep learning: a review},
  author={Ali, Sajid and Masood, Khalid and Riaz, Anas and Saud, Amna},
  booktitle={2022 international conference on business analytics for technology and security (ICBATS)},
  pages={1--7},
  year={2022},
  address = {Dubai, UAE}
}

@article{ancona2023novel,
  title={A novel methodology to disambiguate organization names: an application to {EU Framework Programmes} data},
  author={Ancona, Andrea and Cerqueti, Roy and Vagnani, Gianluca},
  journal={Scientometrics},
  volume={128},
  number={8},
  pages={4447--4474},
  year={2023},
  publisher={Springer}
}

@article{ballreich2021allocation,
  title={Allocation of {National Institutes of Health} funding by disease category in 2008 and 2019},
  author={Ballreich, Jeromie M and Gross, Cary P and Powe, Neil R and Anderson, Gerard F},
  journal={JAMA Network Open},
  volume={4},
  number={1},
  pages={e2034890},
  year={2021}
}

@article{bloch2015size,
  title={The size of research funding: trends and implications},
  author={Bloch, Carter and S{\o}rensen, Mads P},
  journal={Science and Public Policy},
  volume={42},
  number={1},
  pages={30--43},
  year={2015},
  publisher={Oxford University Press}
}

@article{blondel2008fast,
  title={Fast unfolding of communities in large networks},
  author={Blondel, Vincent D and Guillaume, Jean-Loup and Lambiotte, Renaud and Lefebvre, Etienne},
  journal={Journal of Statistical Mechanics: Theory and Experiment},
  volume={2008},
  number={10},
  pages={P10008},
  year={2008}
}

@article{bouarroudj2022named,
  title={Named entity disambiguation in short texts over knowledge graphs},
  author={Bouarroudj, Wissem and Boufaida, Zizette and Bellatreche, Ladjel},
  journal={Knowledge and Information Systems},
  volume={64},
  number={2},
  pages={325--351},
  year={2022},
  publisher={Springer}
}

@inproceedings{bowman2015large,
  title={A large annotated corpus for learning natural language inference},
  author={Bowman, Samuel R and Angeli, Gabor and Potts, Christopher and Manning, Christopher D},
  booktitle={Proceedings of the 2015 conference on empirical methods in natural language processing},
  pages={632--642},
  year={2015},
  address = {Lisbon, Portugal},
}

@article{bromham2016interdisciplinary,
  title={Interdisciplinary research has consistently lower funding success},
  author={Bromham, Lindell and Dinnage, Russell and Hua, Xia},
  journal={Nature},
  volume={534},
  number={7609},
  pages={684--687},
  year={2016},
  publisher={Nature Publishing Group UK London}
}

@misc{cbd2022gbf,
  author       = {{Convention on Biological Diversity}},
  title        = {{Kunming--Montreal Global Biodiversity Framework}},
  year         = {2022},
  url          = {https://www.cbd.int/gbf},
}

@article{dalsgaard2026linking,
  title={Linking Global Science Funding to Research Publications},
  author={Dalsgaard, Jacob Aarup and Silva, Filipi Nascimento and Ai, Jin},
  journal={arXiv preprint arXiv:2603.24147},
  year={2026}
}

@inproceedings{diesner2015impact,
  title={Impact of entity disambiguation errors on social network properties},
  author={Diesner, Jana and Evans, Craig and Kim, Jinseok},
  booktitle={Proceedings of the international {AAAI} conference on web and social media},
  volume={9},
  number={1},
  pages={81--90},
  year={2015},
  address = {Oxford, UK}
}

@article{fortin2013big,
  title={Big science vs. little science: how scientific impact scales with funding},
  author={Fortin, Jean-Michel and Currie, David J},
  journal={PLoS One},
  volume={8},
  number={6},
  pages={e65263},
  year={2013},
  publisher={Public Library of Science}
}

@inproceedings{french2022emerging,
  title={Emerging uses of the research organization registry},
  author={French, Amanda},
  booktitle={Septentrio Conference Series},
  number={1},
  year={2022}
}

@phdthesis{guan2025disambiguating,
  author  = {Guan, Yingjun},
  title   = {Disambiguating academic institution names: a comprehensive study of authority files, linguistic variations, and computational evaluation in {PubMed} affiliations},
  school  = {University of Illinois Urbana-Champaign},
  year    = {2025},
  month   = jul,
  type    = {{Ph.D. Dissertation}},
  address = {Urbana, Illinois},
  url     = {https://hdl.handle.net/2142/130147}
}

@inproceedings{hadsell2006dimensionality,
  title={Dimensionality reduction by learning an invariant mapping},
  author={Hadsell, Raia and Chopra, Sumit and LeCun, Yann},
  booktitle={2006 IEEE computer society conference on computer vision and pattern recognition (CVPR'06)},
  volume={2},
  pages={1735--1742},
  year={2006},
  address = {New York, USA}
}

@article{henderson2017efficient,
  title={Efficient natural language response suggestion for smart reply},
  author={Henderson, Matthew and Al-Rfou, Rami and Strope, Brian and Sung, Yun-Hsuan and Luk{\'a}cs, L{\'a}szl{\'o} and Guo, Ruiqi and Kumar, Sanjiv and Miklos, Balint and Kurzweil, Ray},
  journal={arXiv preprint arXiv:1705.00652},
  year={2017}
}

@article{hendricks2020crossref,
  title={Crossref: The sustainable source of community-owned scholarly metadata},
  author={Hendricks, Ginny and Tkaczyk, Dominika and Lin, Jennifer and Feeney, Patricia},
  journal={Quantitative Science Studies},
  volume={1},
  number={1},
  pages={414--427},
  year={2020},
  publisher={MIT Press One Rogers Street, Cambridge, MA 02142-1209, USA journals-info~…}
}

@article{huang2014institution,
  title={Institution name disambiguation for research assessment},
  author={Huang, Shuiqing and Yang, Bo and Yan, Sulan and Rousseau, Ronald},
  journal={Scientometrics},
  volume={99},
  number={3},
  pages={823--838},
  year={2014},
  publisher={Springer}
}

@article{jacob2011impact,
  title={The impact of research grant funding on scientific productivity},
  author={Jacob, Brian A and Lefgren, Lars},
  journal={Journal of Public Economics},
  volume={95},
  number={9-10},
  pages={1168--1177},
  year={2011},
  publisher={Elsevier}
}

@article{jiang2011affiliation,
  title={Affiliation disambiguation for constructing semantic digital libraries},
  author={Jiang, Yong and Zheng, Hai-Tao and Wang, Xinmin and Lu, Binggan and Wu, Kaihua},
  journal={Journal of the American Society for Information Science and Technology},
  volume={62},
  number={6},
  pages={1029--1041},
  year={2011},
  publisher={Wiley Online Library}
}

@article{jonnalagadda2010nemo,
  title={{NEMO}: Extraction and normalization of organization names from {PubMed} affiliation strings},
  author={Jonnalagadda, Siddhartha and Topham, Philip},
  journal={Journal of Biomedical Discovery and Collaboration},
  volume={5},
  pages={50},
  year={2010}
}

@inproceedings{joulin2017bag,
  title={Bag of tricks for efficient text classification},
  author={Joulin, Armand and Grave, Edouard and Bojanowski, Piotr and Mikolov, Tom{\'a}{\v{s}}},
  booktitle={Proceedings of the 15th conference of the European chapter of the association for computational linguistics},
  pages={427--431},
  year={2017},
  address = {Valencia, Spain},
}

@article{kim2015effect,
  title={The effect of data pre-processing on understanding the evolution of collaboration networks},
  author={Kim, Jinseok and Diesner, Jana},
  journal={Journal of Informetrics},
  volume={9},
  number={1},
  pages={226--236},
  year={2015},
  publisher={Elsevier}
}

@article{kim2016distortive,
  title={Distortive effects of initial-based name disambiguation on measurements of large-scale coauthorship networks},
  author={Kim, Jinseok and Diesner, Jana},
  journal={Journal of the Association for Information Science and Technology},
  volume={67},
  number={6},
  pages={1446--1461},
  year={2016},
  publisher={Wiley Online Library}
}

@inproceedings{kim2014name,
  title={Why name ambiguity resolution matters for scholarly big data research},
  author={Kim, Jinseok and Diesner, Jana and Kim, Heejun and Aleyasen, Amirhossein and Kim, Hwan-Min},
  booktitle={2014 {IEEE} international conference on big data},
  pages={1--6},
  year={2014},
  address = {Washington, D.C., USA}
}

@article{kokol2018discrepancies,
  title={Discrepancies among {Scopus, Web of Science, and PubMed} coverage of funding information in medical journal articles},
  author={Kokol, Peter and Vo{\v{s}}ner, Helena Bla{\v{z}}un},
  journal={Journal of the Medical Library Association},
  volume={106},
  number={1},
  pages={81},
  year={2018}
}

@article{lammey2020solutions,
  title={Solutions for identification problems: a look at the Research Organization Registry},
  author={Lammey, Rachael},
  journal={Science Editing},
  volume={7},
  number={1},
  pages={65--69},
  year={2020},
  publisher={Korean Council of Science Editors}
}

@inproceedings{liu2025not,
  title={Not-just-scaling laws: towards a better understanding of the downstream impact of language model design decisions},
  author={Liu, Emmy and Bertsch, Amanda and Sutawika, Lintang and Tjuatja, Lindia and Fernandes, Patrick and Marinov, Lara and Chen, Michael and Singhal, Shreya and Lawrence, Carolin and Raghunathan, Aditi and others},
  booktitle={Proceedings of the 2025 conference on empirical methods in natural language processing},
  pages={16407--16438},
  year={2025},
  address = {Suzhou, China}
}

@article{liu2020accuracy,
  title={Accuracy of funding information in {Scopus}: a comparative case study},
  author={Liu, Weishu},
  journal={Scientometrics},
  volume={124},
  number={1},
  pages={803--811},
  year={2020},
  publisher={Springer}
}

@article{mishra2018self,
  title={Self-citation is the hallmark of productive authors, of any gender},
  author={Mishra, Shubhanshu and Fegley, Brent D and Diesner, Jana and Torvik, Vetle I},
  journal={PLoS One},
  volume={13},
  number={9},
  pages={e0195773},
  year={2018},
  publisher={Public Library of Science San Francisco, CA USA}
}

@article{morillo2013towards,
  title={Towards the automation of address identification},
  author={Morillo, Fernanda and Aparicio, Javier and Gonz{\'a}lez-Albo, Borja and Moreno, Luz},
  journal={Scientometrics},
  volume={94},
  number={1},
  pages={207--224},
  year={2013},
  publisher={Springer}
}

@book{oecd2024biodiversity,
  author    = {{OECD}},
  title     = {Biodiversity and development finance 2015--2022: contributing to Target 19 of the Kunming--Montreal Global Biodiversity Framework},
  year      = {2024},
  publisher = {OECD Publishing},
  address   = {Paris}
}

@article{onodera2011method,
  title={A method for eliminating articles by homonymous authors from the large number of articles retrieved by author search},
  author={Onodera, Natsuo and Iwasawa, Mariko and Midorikawa, Nobuyuki and Yoshikane, Fuyuki and Amano, Kou and Ootani, Yutaka and Kodama, Tadashi and Kiyama, Yasuhiko and Tsunoda, Hiroyuki and Yamazaki, Shizuka},
  journal={Journal of the American Society for Information Science and Technology},
  volume={62},
  number={4},
  pages={677--690},
  year={2011},
  publisher={Wiley Online Library}
}

@article{paul2016characterization,
  title={Characterization, description, and considerations for the use of funding acknowledgement data in {Web of Science}},
  author={Paul-Hus, Ad{\`e}le and Desrochers, Nadine and Costas, Rodrigo},
  journal={Scientometrics},
  volume={108},
  number={1},
  pages={167--182},
  year={2016},
  publisher={Springer}
}

@article{pranckute2021web,
  title={{Web of Science (WoS) and Scopus}: the titans of bibliographic information in today’s academic world},
  author={Pranckut{\.e}, Raminta},
  journal={Publications},
  volume={9},
  number={1},
  pages={12},
  year={2021},
  publisher={MDPI}
}

@article{rabovsky2014higher,
  title={Higher education and congressional influence on administrative decisions: an examination of {NSF} and {NIH} research grant funding to four-year universities},
  author={Rabovsky, Thomas M and Ellis, William Curtis},
  journal={Social Science Quarterly},
  volume={95},
  number={3},
  pages={740--759},
  year={2014},
  publisher={Wiley Online Library}
}

@inproceedings{reimers-2019-sentence-bert,
  title={Sentence-{BERT}: sentence embeddings using {Siamese BERT}-networks},
  author={Reimers, Nils and Gurevych, Iryna},
  booktitle={Proceedings of the 2019 conference on empirical methods in natural language processing and the 9th international joint conference on natural language processing (EMNLP-IJCNLP)},
  pages={3982--3992},
  year={2019},
  address = {Hong Kong, China}
}

@inproceedings{roberts2020much,
  title={How much knowledge can you pack into the parameters of a language model?},
  author={Roberts, Adam and Raffel, Colin and Shazeer, Noam},
  booktitle={Proceedings of the 2020 conference on empirical methods in natural language processing (EMNLP)},
  pages={5418--5426},
  year={2020}
}

@article{sanyal2021review,
  title={A review of author name disambiguation techniques for the {PubMed} bibliographic database},
  author={Sanyal, Debarshi Kumar and Bhowmick, Plaban Kumar and Das, Partha Pratim},
  journal={Journal of Information Science},
  volume={47},
  number={2},
  pages={227--254},
  year={2021},
  publisher={SAGE Publications Sage UK: London, England}
}

@article{shao2020elad,
  title={{ELAD}: an entity linking based affiliation disambiguation framework},
  author={Shao, Zhou and Cao, Xiangying and Yuan, Sha and Wang, Yongli},
  journal={IEEE Access},
  volume={8},
  pages={70519--70526},
  year={2020},
  publisher={IEEE}
}

@inproceedings{song2020research,
  title={Research on organization name matching based on word vector},
  author={Song, KaiLi and Li, YunLing and Yao, LuLu and Wang, Yuan},
  booktitle={Journal of Physics: Conference Series},
  volume={1684},
  number={1},
  pages={012085},
  year={2020},
  organization={IOP Publishing}
}

@article{sun2017using,
  title={Using an ontology-based approach to handle author affiliations in a large biomedical citation database},
  author={Sun, Haixia and Li, Junlian and Wu, Yingjie and Wang, Lei and Fung, Kin Wah},
  journal={Studies in Health Technology and Informatics},
  volume={245},
  pages={1338},
  year={2017}
}

@article{vera2025embeddinggemma,
  title={{EmbeddingGemma}: powerful and lightweight text representations},
  author={Vera, Henrique Schechter and Dua, Sahil and Zhang, Biao and Salz, Daniel and Mullins, Ryan and Panyam, Sindhu Raghuram and Smoot, Sara and Naim, Iftekhar and Zou, Joe and Chen, Feiyang and others},
  journal={arXiv preprint arXiv:2509.20354},
  year={2025}
}

@article{waldron2017reductions,
  title={Reductions in global biodiversity loss predicted from conservation spending},
  author={Waldron, Anthony and Miller, Daniel C and Redding, Dave and Mooers, Arne and Kuhn, Tyler S and Nibbelink, Nate and Roberts, J Timmons and Tobias, Joseph A and Gittleman, John L},
  journal={Nature},
  volume={551},
  number={7680},
  pages={364--367},
  year={2017},
  publisher={Nature Publishing Group}
}

@article{waldron2013targeting,
  title={Targeting global conservation funding to limit immediate biodiversity declines},
  author={Waldron, Anthony and Mooers, Arne O and Miller, Daniel C and Nibbelink, Nate and Redding, David and Kuhn, Tyler S and Roberts, J Timmons and Gittleman, John L},
  journal={Proceedings of the National Academy of Sciences},
  volume={110},
  number={29},
  pages={12144--12148},
  year={2013},
  publisher={National Academy of Sciences}
}

@article{whitley2018impact,
  title={The impact of changing funding and authority relationships on scientific innovations},
  author={Whitley, Richard and Gl{\"a}ser, Jochen and Laudel, Grit},
  journal={Minerva},
  volume={56},
  pages={109--134},
  year={2018},
  publisher={Springer}
}

@inproceedings{williams2018broad,
  title={A broad-coverage challenge corpus for sentence understanding through inference},
  author={Williams, Adina and Nangia, Nikita and Bowman, Samuel R},
  booktitle={Proceedings of the 2018 conference of the North American chapter of the association for computational linguistics: human language technologies},
  pages={1112--1122},
  year={2018},
  address = {New Orleans, Louisiana}
}

@article{xie2014undemocracy,
  title={“{Undemocracy}”: inequalities in science},
  author={Xie, Yu},
  journal={Science},
  volume={344},
  number={6186},
  pages={809--810},
  year={2014},
  publisher={American Association for the Advancement of Science}
}

@article{yang2024does,
  author={Yang, Zhen and Ding, Ming and Huang, Tinglin and Cen, Yukuo and Song, Junshuai and Xu, Bin and Dong, Yuxiao and Tang, Jie},
  journal={IEEE Transactions on Pattern Analysis and Machine Intelligence}, 
  title={Does Negative Sampling Matter? a Review With Insights Into its Theory and Applications}, 
  year={2024},
  volume={46},
  number={8},
  pages={5692-5711}
}

@inproceedings{zhang2024scaling,
  title={When scaling meets {LLM} finetuning: the effect of data, model and finetuning method},
  author={Zhang, Biao and Liu, Zhongtao and Cherry, Colin and Firat, Orhan},
  booktitle={International conference on learning representations},
  volume={2024},
  pages={44694--44713},
  year={2024},
  address = {Vienna, Austria}
}

@article{zhang2023lagos,
  title={{LAGOS-AND}: a large gold standard dataset for scholarly author name disambiguation},
  author={Zhang, Li and Lu, Wei and Yang, Jinqing},
  journal={Journal of the Association for Information Science and Technology},
  volume={74},
  number={2},
  pages={168--185},
  year={2023},
  publisher={Wiley Online Library}
}

@article{zhang2025qwen3,
  title={Qwen3 embedding: advancing text embedding and reranking through foundation models},
  author={Zhang, Yanzhao and Li, Mingxin and Long, Dingkun and Zhang, Xin and Lin, Huan and Yang, Baosong and Xie, Pengjun and Yang, An and Liu, Dayiheng and Lin, Junyang and others},
  journal={arXiv preprint arXiv:2506.05176},
  year={2025}
}

@article{zhi2016,
author = {Zhi, Qiang and Meng, Tianguang},
title = {Funding allocation, inequality, and scientific research output: an empirical study based on the life science sector of {Natural Science Foundation of China}},
year = {2016},
publisher = {Springer-Verlag},
address = {Berlin, Heidelberg},
volume = {106},
number = {2},
journal = {Scientometrics},
pages = {603–628},
numpages = {26}
}

@article{zhou2020depth,
  title={An in-depth analysis of government funding and international collaboration in scientific research},
  author={Zhou, Ping and Cai, Xiaojing and Lyu, Xiaozan},
  journal={Scientometrics},
  volume={125},
  pages={1331--1347},
  year={2020},
  publisher={Springer}
}
\end{document}